%% file: main.tex
\documentclass[sigconf]{acmart}
\AtBeginDocument{%
  }

\setcopyright{acmlicensed}
\copyrightyear{2025}
\acmYear{2025}
\acmDOI{10.1145/3711896.3736947}
\acmConference[KDD '25] {Proceedings of the 31st ACM SIGKDD Conference on Knowledge Discovery and Data Mining V.2}{August 3--7, 2025}{Toronto, ON, Canada.}
\acmBooktitle{Proceedings of the 31st ACM SIGKDD Conference on Knowledge Discovery and Data Mining V.2 (KDD '25), August 3--7, 2025, Toronto, ON, Canada}
\acmISBN{979-8-4007-1454-2/25/08} 

\usepackage[capitalize,noabbrev]{cleveref}
\usepackage{algorithm}
\usepackage[noend]{algorithmic}
\usepackage{multirow}

\AtEndPreamble{%
\theoremstyle{acmplain}
\newtheorem{assumption}[theorem]{Assumption}
\theoremstyle{acmdefinition}
\newtheorem{remark}[theorem]{Remark}
}

\DeclareMathOperator{\CE}{CE}
\DeclareMathOperator*{\E}{\mathbb{E}}
\DeclareMathOperator*{\argmax}{argmax}

\DeclareMathOperator{\score}{score}
\DeclareMathOperator{\parent}{parent}
\newcommand{\gradcum}{\operatorname{grad}_{\text{cum}}}

\newcommand{\R}{\mathbb{R}}
\newcommand{\Enn}{\mathcal{N}}
\newcommand{\1}{\mathbf{1}}

\newcommand{\partials}[2][]{\frac{\partial #1}{\partial #2}}
\newcommand{\norm}[1]{\left\lVert#1\right\rVert}

\newcommand{\jesse}[1] {}
\newcommand{\yusu}[1] {}
\newcommand{\yusuedit}[1] {}

\begin{document}

\title{Explaining GNN Explanations with Edge Gradients}

\author{Jesse He}
\email{jeh020@ucsd.edu}
\affiliation{%
  \department{Halıcıoğlu Data Science Institute}
  \institution{Universiy of California, San Diego}
  \city{San Diego}
  \state{California}
  \country{USA}
}
\author{Akbar Rafiey}
\email{ar9530@nyu.edu}
\affiliation{%
  \department{Computer Science and Engineering}
  \institution{New York University}
  \city{Brooklyn}
  \state{New York}
  \country{USA}
}
\author{Gal Mishne}
\email{gmishne@ucsd.edu}
\affiliation{%
  \department{Halıcıoğlu Data Science Institute}
  \institution{Universiy of California, San Diego}
  \city{San Diego}
  \state{California}
  \country{USA}
}
\author{Yusu Wang}
\email{yusuwang@ucsd.edu}
\affiliation{%
  \department{Halıcıoğlu Data Science Institute}
  \institution{Universiy of California, San Diego}
  \city{San Diego}
  \state{California}
  \country{USA}
}

\input{sections/abstract}



\received{10 February 2025}

\maketitle
\newcommand\kddavailabilityurl{https://doi.org/10.5281/zenodo.15485808}

\ifdefempty{\kddavailabilityurl}{}{
\begingroup\small\noindent\raggedright\textbf{KDD Availability Link:}\\
The source code of this paper has been made publicly available at \url{\kddavailabilityurl}.
\endgroup
}

\input{sections/introduction}
\input{sections/related-works}
\input{sections/preliminaries}
\input{sections/grad-and-perturb}
\input{sections/other-methods}
\input{sections/experiments}
\input{sections/conclusion}

\begin{acks}
This work was partially supported by NSF awards CCF-2217058, 2112665, 2403452, and 2310411. J.H. was also supported by a 2024 Qualcomm Innovation Fellowship.
\end{acks}

\bibliographystyle{ACM-Reference-Format}
\bibliography{references}

\appendix

\input{appendix/proofs}
\begin{table*}
    \centering
    \input{appendix/accuracy_table}
    \caption{Accuracy (node classification) or AUROC (graph classification) of trained models on each dataset in \cref{subsec:real-data}}
    \label{tab:model-accuracy}
\end{table*}
\input{appendix/regularization}
\input{appendix/training}

\end{document}

%% file: sections/abstract.tex
\begin{abstract}
In recent years, the remarkable success of graph neural networks (GNNs) on graph-structured data has prompted a surge of methods for explaining GNN predictions. However, the state-of-the-art for GNN explainability remains in flux. Different comparisons find mixed results for different methods, with many explainers struggling on more complex GNN architectures and tasks. This presents an urgent need for a more careful theoretical analysis of competing GNN explanation methods. In this work we take a closer look at GNN explanations in two different settings: \emph{input-level} explanations, which produce explanatory subgraphs of the input graph, and \emph{layerwise} explanations, which produce explanatory subgraphs of the computation graph. We establish the first theoretical connections between the popular perturbation-based and classical gradient-based methods, as well as point out connections between other recently proposed methods. At the input level, we demonstrate conditions under which GNNExplainer can be approximated by a simple heuristic based on the sign of the edge gradients. In the layerwise setting, we point out that edge gradients are equivalent to occlusion search for linear GNNs. Finally, we demonstrate how our theoretical results manifest in practice with experiments on both synthetic and real datasets.
\end{abstract}

%% file: sections/introduction.tex
\section{Introduction}

In response to the emergence of Graph Neural Networks (GNNs) for machine learning on graph-structured data, there has been a recent surge of methods aimed to \emph{explain} GNNs. While explainability has been studied extensively in other domains \cite{bach2015pixel, ribeiro2016should, lundberg2017unified, selveraju2017gradcam, sundararajan2017axiomatic, ancona2018, adebayo2018sanity}, only recently have methods been proposed to deal with GNNs and the unique challenges posed by graph-structured input. In addition to the input node features, GNNs also use the given graph structure to make predictions. Therefore, explainability methods are tasked with identifying important \emph{subgraphs}. In this work we focus on \emph{post-hoc} explainers, which explain the predictions of a fixed, trained GNN. We also focus on \emph{instance-based} explanation, where each prediction is explained separately, irrespective of the model's behavior on other samples. Finally, we focus on explanations in terms of the graph topology rather than the node features.

Although a number of empirical comparisons have been performed between various methods \cite{faber2021, pmlr-v198-amara22a, agarwal2023evaluating, longa2025explainingtheexplainers}, the state of the art remains unclear. The efficacy of different methods is mixed across comparisons on different tasks with different GNN models. For example, saliency maps are consistently a strong baseline, often outperforming more sophisticated methods despite their simplicity \cite{pmlr-v198-amara22a}. In addition, while most explainability methods are evaluated on the straightforward Graph Convolutional Network (GCN) \cite{kipf2017semisupervisedclassificationgraphconvolutional}, \citet{longa2025explainingtheexplainers} find that other popular architectures such as Graph Isomorphism Networks (GIN) \cite{xu2018powerful} are substantially more difficult for explainers. Moreover, some methods are brittle on real data despite strong performance on synthetic datasets with clear ground-truth explanations. Alarmingly, Agarwal et al.~\cite{pmlr-v151-agarwal22b} find that on some datasets, existing methods may struggle to outperform even randomly selecting edges in terms of unfaithfulness.

These findings present an urgent need for a clearer theoretical understanding of GNN explainers. While some gradient and decomposition-based methods have been unified in other settings \cite{ancona2018}, the novelty of many GNN-specific methods means that there is very little unified theory.
Our analysis helps both to clarify the state of explainability in GNNs and to suggest additional avenues for future research.
Our contributions are as follows:
\begin{enumerate}
    \item We identify conditions under which GNNExplainer produces explanations which are similar to a simple gradient-based heuristic (\cref{subsec:one-layer}).
    \item Using a decomposition of the computation graph into paths, we show that Layerwise Occlusion and Layerwise Gradients are equivalent for linear GNNs (\cref{subsec:layerwise}).
    \item Leveraging this path decomposition, we also obtain a new interpretation of AMP-ave \cite{pmlr-v202-xiong23b}, a recent algorithm to approximate relevant paths for a GNN. We use this interpretation to describe a simple algorithm that performs an exact computation. We also draw a connection to GOAt \cite{lu2024goat}, another recently proposed decomposition method (\cref{sec:connections-other}).
    \item We validate our analysis through extensive experiments on synthetic and real data. We also demonstrate that our simple algorithm for computing relevant paths outperforms AMP-ave (\cref{sec:experiments}).
\end{enumerate}
Taken together, our results show that explanability for GNNs remains closely tied to the behavior of gradient-based methods, which are simple and easy to compute. As the need for explainable GNNs continues to grow, we hope this work will provide a more unified perspective on the design space of possible explanation methods.

%% file: sections/related-works.tex
\section{Related Works}
\label{sec:related-works}

Numerous explainability methods have been proposed for GNNs, and they are often often grouped into several distinct categories:
\emph{Gradient-based} methods such as saliency maps \cite{pope2019explainability}, Gradient$\odot$Input \cite{pmlr-v70-shrikumar17a}, or Integrated Gradients \cite{sundararajan2017axiomatic} use backpropagation to compute a function of the network gradients with respect to the node input features or the adjacency matrix.
\emph{Decomposition-based methods} attempt to decompose the output in terms of the inputs. They may also use backpropagation to compute attributions, as in CAM \cite{zhou2016learning}, LRP \cite{bach2015pixel,schnake2021higher}, or DeepLIFT \cite{pmlr-v70-shrikumar17a}, or they may use other strategies based on knowledge of the network like DEGREE \cite{feng2022degree} and GOAt \cite{lu2024goat}. \emph{Perturbation-based} methods study how the network prediction changes as the input changes. The most simple is Occlusion search \cite{zeiler2014visualizing}, but recently more sophisticated methods have become popular for GNNs such as GNNExplainer \cite{ying2019gnnexplainer}, PGExplainer \cite{luo2020pgexplainer}, and GraphMask \cite{schlichtkrull2022interpretinggraphneuralnetworks}. \emph{Search-based} methods such as SubgraphX \cite{pmlr-v139-yuan21c}, sGNN-LRP \cite{xiong2022efficient}, and EiG-Search \cite{pmlr-v235-lu24g} attempt to build an explanatory subgraph by performing a heuristic search over the space of possible subgraphs. \emph{Surrogate-based} methods like GraphLIME \cite{huang20233graphlime} and PGM-Explainer \cite{vu2020pgmexplainer} train a simpler interpretable model which behaves similarly to the target GNN.

The plethora of competing methods has prompted several empirical evaluations, including \cite{faber2021, pmlr-v198-amara22a, agarwal2023evaluating, longa2025explainingtheexplainers}.
However, despite these empirical comparisons, their theoretical properties have only recently begun to be explored. \citet{pmlr-v151-agarwal22b} establish theoretical bounds for the faithfulness, stability, and fairness of various explainers in terms of the Lipschitz constants of the original GNN or its activation functions. \citet{fang2023robustness} study the robustness of GNN explainers to adversarial perturbations. Similarly, \citet{li2024fragile} find that GNN explanations are fragile when faced with adversarial attacks. Unlike our work, however, these works do not seek to establish similarities between existing GNN explanation methods.

Outside of the graph domain, more work has been done to probe different explainability methods. \citet{ancona2018} unify gradient-based methods with the decomposition methods LRP \cite{bach2015pixel} and Deep-LIFT \cite{pmlr-v70-shrikumar17a}. \citet{adebayo2018sanity}, show that some gradient-based and decomposition methods in computer vision approximate edge detectors when applied to convolutional neural networks, implying that qualitative visual assessment of explanations can be misleading. Our contribution builds on these works by unifying explainability methods in the novel setting of GNNs.

%% file: sections/preliminaries.tex
\section{Preliminaries}

\subsection{Graph Neural Networks}

We begin by introducing some preliminary notions and notation. Throughout, let $G = (V,E)$ be an unweighted (possibly directed) graph with node features $x_v \in \R^p$ for each $v \in V$. We will treat undirected graphs as bidirected, with separate edges $(u,v)$ and $(v,u)$. We will denote by $\Enn(v)$ the neighborhood of a node $v$ and by $\Enn^{(k)}(v)$ the $k$-hop neighborhood of $v$ (that is, all nodes within $k$ hops of $v$). Recall that a message-passing graph neural network (GNN) consists of $L$ layers, in which each node aggregates its neighbors' embeddings and updates its own. 
Formally, in each layer $\ell$ the node embedding $x_v^{(\ell)}$ of node $v$ is given by
\begin{equation}
    x_v^{(\ell)} = f^{(\ell)}_{\text{Update}}\left(
    f^{(\ell)}_{\text{Aggregate}}(\{x_u^{(\ell-1)} : u \in \Enn(v)\}).
    \right)
\end{equation}
We will denote the output of a GNN $\Phi$ at a node $v \in G$ by $\Phi(v;G)$.
A useful perspective is to consider the \emph{computation graph} of a node $v$ with respect to an $L$-layer GNN:
\begin{definition}
    The \emph({$L$-layer) computation graph} $G_c(v)$ of a node $v$ in a base graph $G$ is the directed graph with vertex set $V(G_c(v)) = \bigsqcup_{\ell = 0}^L\Enn^{(\ell)}(v)$ and edges $(i^{(\ell-1)},j^{(\ell)})$ for each $(i,j) \in E(G)$ and each $j \in \Enn^{(L-\ell)}(v)$. We refer to the set of edges $(i^{(\ell-1)},j^{(\ell)})$ in layer $\ell$ by $E^{(\ell)}(G_c(v))$. We may write $G_c$ when $v$ is fixed.
\end{definition}

Essentially, the computation graph $G_c(v)$ encodes the flow of information from input features into the final prediction for a specific node $v$. An important property of the computation graph $G_c$ is that it is directed and acyclic. Ordering each node $j^{(\ell)}$ of $G_c$ by its layer $\ell$ defines a topological ordering of $G_c$. We will describe how to take advantage of this property in \cref{sec:connections-other}.

\subsection{Explanation Methods}

The goal of explanation methods for GNNs is to identify an important subgraph of the input graph which explains why the network classifies a certain node with a particular label.
In other words, given a GNN $\Phi$ and its prediction $Y_v$ at a node $v$ in a graph $G$, we wish to identify a subgraph $G_S$ of $G$ which explains why $\Phi$  predicts $Y_v$. We will focus on selecting subgraphs by identifying important edges, regarding node features to be of equal importance. That is, we consider attribution methods of the form $h : E(G) \to \R$.
One can also generate explanations in the computation graph $G_c$, giving attributions $h_c : E(G_c) \to \R$. By producing explanations in $G_c$, we can identify important paths that lead to the final prediction at node $v$.
In this section we describe the explanation methods we will study in \cref{sec:connecting-grad-perturb}.

\paragraph{Edge Gradients \cite{pope2019explainability}}
In more classical neural models such as multi-layer perceptions or convolutional networks\yusu{what are "other domains?"}\jesse{clarified to computer vision}\yusu{How about "in the setting of more classical learnign models such as MLP or CNN", to point out that graph leanring setting is different?}\jesse{That makes sense, thanks}, multiple explainability methods rely on computing the gradient of the output with respect to the input features \cite{pmlr-v70-shrikumar17a, sundararajan2017axiomatic, selveraju2017gradcam}. In the GNN setting, one can also compute gradients with respect to the graph structure by treating the graph adjacency matrix as an input feature. Each edge $(i,j)$ with weight $\omega_{ij}$ is given the attribution
\begin{equation}
    \operatorname{\textsc{Grad}}_v(i,j) = \partials{\omega_{ij}} \Phi(v;G).
\end{equation}
Since we consider unweighted graphs, we take $\omega_{ij} = 1$ for edges $(i,j)$ in the original input graph. However, our analysis can be extended to weighted graphs so long as $\omega_{ij} \geq 0$.

\paragraph{Occlusion \cite{zeiler2014visualizing}}
A simple perturbation-based baseline for edge attribution is \emph{occlusion}, which studies how $\Phi(v;G)$ changes when an edge is removed. An edge $(i,j)$ is given the attribution
\begin{equation}
    \operatorname{\textsc{Occ}}_v(i,j) = \Phi(v;G) - \Phi(v; G \setminus (i,j))
\end{equation}
where $\Phi(v; G \setminus (i,j))$ is the output of the network on the perturbed graph with edge $(i,j)$ removed.

\paragraph{GNNExplainer \cite{ying2019gnnexplainer}}
GNNExplainer is one of the earliest and most influential explainability methods designed for GNNs. Its mutual information-based objective has been adopted by \cite{luo2020pgexplainer, wu2020gib, miao2022interpretable}. GNNExplainer optimizes a subgraph $G_S \subset G$ which maximizes the mutual information $I(Y_v; G_S)$ between the label $Y_v$ and the subgraph $G_S$. To optimize $G_S$, GNNExplainer relaxes the problem to produce a \emph{soft mask} $\Omega$ on $G$. Then $G_S$ is treated as a weighted graph with weight matrix $\Omega = (\omega_{ij})$, which is optimized via gradient descent. The entries $\omega_{ij}$ serve as the attributions for explanation.

In practice, optimizing the mutual information directly is difficult, so the objective is relaxed to minimizing the cross-entropy between the original prediction and the output on the masked graph $G_S$. Moreover, most implementations of GNNExplainer include regularization terms to encourage desirable properties in the final explanation. The most popular, such as in PyTorch Geometric \cite{pyg}, include a subgraph size constraint $\alpha \norm{\Omega}_1$ to promote sparsity and an entropy constraint $\beta H(\Omega)$ to further encourage binary-valued mask entries:
\begin{equation}
    H(\Omega) = - \frac{1}{|E|} \sum_{(i,j) \in E}
    (1-\omega_{ij})\log(1-\omega_{ij}) + \omega_{ij}\log(\omega_{ij})
\end{equation}
where $\alpha, \beta$ are hyperparameters.
Thus, GNNExplainer's objective is reformulated as
\begin{equation}
    \min_{\Omega} \CE(Y_v, \Phi(v; G_S(\Omega))) + \alpha \norm{\Omega}_1 + \beta H(\Omega)
    \label{eq:gnnexplainer}
\end{equation}
which is minimized using gradient descent.

%% file: sections/grad-and-perturb.tex
\section{Connecting Gradients and Perturbation-based Explanations}
\label{sec:connecting-grad-perturb}

In this section, we draw theoretical connections between edge gradients and perturbation-based methods.
We will consider the following explanation settings:
\begin{enumerate}
    \item Explanations on the \emph{input graph}. Here, we consider the problem of selecting $G_S$ as a subgraph of the input graph $G$, as in \cite{ying2019gnnexplainer}. As we will see, GNNExplainer's attributions are governed by the sign of the edge gradients.
    \item Explanations on the \emph{computation graph}. Here, we will allow for different attributions to be given to each ``copy'' of an edge in the computation graph, essentially producing a subgraph for each layer of the network as in \cite{schlichtkrull2022interpretinggraphneuralnetworks}. We will see that layerwise edge gradients are equivalent to layerwise occlusion for linear networks.
\end{enumerate}
For simplicity, we will adopt the following setup: suppose $\Phi$ is a sum-aggregated message-passing network. We will omit bias terms, so each intermediate layer $\ell = 1, \dots, L-1$ is given by
\begin{equation}
    x_v^{(\ell)} = \varphi^{(\ell)} \left( \sum_{u \in \Enn(v)} W^{(\ell)}x^{(\ell-1)}_{u}\right)
    \label{eq:gnn-layer}
\end{equation}
where $\varphi^{(\ell)}$ is a (possibly nonlinear) function.
The final layer has no nonlinearity:
\begin{equation}
    \Phi(v;G) = z_v^{(L)} = \sum_{u \in \Enn(v)} W^{(L)}x^{(L-1)}_{u}.
\end{equation}
 Suppose also that the task is binary node classification with labels in $\{+,-\}$ where the probability of node $v$ being class $+$ is given by $\sigma(\Phi(v;G))$. That is, we apply the sigmoid $\sigma$ to the output $\Phi(v;G) \in \R$. For brevity, we may also write $z_v \coloneqq \Phi(v;G)$. Although we consider only this simplified setting, our experiments show that this behavior is present in more complex models, multiclass classification tasks, and graph classification. Before we consider the input-level and layerwise settings separately, however, we will begin with the one-layer setting, where they coincide.

\subsection{One-Layer Networks}
\label{subsec:one-layer}

Suppose $\Phi$ is a sum-aggregated linear GNN with $L = 1$ performing binary classification. As above, the probability of class $+$ is estimated with $P_\Phi(Y_v = +) = \sigma(\Phi(v;G))$.
Without loss of generality, we may assume $Y_v = +$, i.e. $\Phi(v;G) > 0$. (Otherwise, we may replace $\Phi$ with $-\Phi$.)
In the one-layer case, we have only one weight matrix $W$ (without a bias term).
Then the underlying graph of $G_c(v)$ is simply $\mathcal{N}(v)$ and, noticing that $\partials[z_v]{x_u} = W$,
\begin{equation}
    \Phi(v;G) = \sum_{u \in \Enn(v)} \partials[z_v]{x_u} \cdot x_u.
    \label{eq:one-layer-gi}
\end{equation}

In this simplified one-layer case, there is a clear connection between gradient-based methods and GNNExplainer, which we prove in \cref{proof:1-layer}. 
\begin{proposition}
    \label{prop:1-layer}
    For a GNN $\Phi$ with one linear layer, the mutual information $I(Y_v;G_S)$ is maximized when $G_S$ consists of each node $u$ and edge $(u,v)$ for which $\partials[z_v]{x_u} \cdot x_u > 0$.
\end{proposition}
In particular, \cref{prop:1-layer} tells us that in the case of a linear network with a single layer, GNNExplainer is equivalent to simply taking the sign of the edge gradients. Moreover, as shown in \cite{ancona2018}, different gradient-based attributions like Gradient$\odot$Input \cite{pmlr-v70-shrikumar17a}, 0-LRP \cite{bach2015pixel}, and Integrated Gradients \cite{sundararajan2017axiomatic} are all equivalent to each other in the one-layer setting, as well as to Occlusion.

\subsection{Input-level Explanations}
\label{subsec:input-level}

We consider the most common setting, where we wish to find an explanatory subgraph of $G$ for the prediction of the network $\Phi$ at node $v$. Here the one-layer case will be illustrative in describing the behavior of GNNExplainer: we point out that because GNNExplainer trains an edge mask via gradient descent, its optimization is governed by the edge gradients. Again, consider binary classification with labels $Y_v \in \{+, -\}$, and assume $Y_v = +$.
\begin{proposition}
    Suppose that for all $\omega_{ij} \in [0,1]$, we have $\partials[z_v]{\omega_{ij}} > 0$. Then
    \begin{equation}
        \argmax_{\omega_{ij}^* \in [0,1]} I(Y_v \mid \omega_{ij} = \omega_{ij}^*) = 1.
    \end{equation}
    Similarly, if $\partials[z_v]{\omega_{ij}} < 0$ for all values of $\omega_{ij}$ in $[0,1]$,
    \begin{equation}
        \argmax_{\omega_{ij}^* \in [0,1]} I(Y_v \mid \omega_{ij} = \omega_{ij}^*) = 0.
    \end{equation}
    \label{prop:no-grad-flip}
\end{proposition}
\cref{prop:no-grad-flip} has a straightforward interpretation: GNNExplainer follows the sign of the edge gradients. This explains, for example, the observation in \cite{faber2021} that GNNExplainer fails to distinguish negative evidence from irrelevant evidence. However, it also implies that GNNExplainer can be used to find negative evidence by changing the target label, as we will see in \cref{subsec:neg-evidence}.
While the condition that $\partials[z_v]{\omega_{ij}}$ is the same sign for all values of $\omega_{ij}$ is quite strong, it holds surprisingly often in practice, as we will see in \cref{sec:experiments}. We discuss the effects of the regularization terms $\alpha\norm{\Omega}_1$ and $\beta H(\Omega)$ in \cref{appdx:regularization}. 

\subsection{Layerwise Explanations}
\label{subsec:layerwise}

Here, we consider the problem of selecting an explanatory subgraph of the computation graph as considered by \cite{schlichtkrull2022interpretinggraphneuralnetworks, schnake2021higher, pmlr-v202-xiong23b}. This approach can identify important paths in $G$ which may not be apparent from input-level explanations. Attach to each edge $\left(i^{(\ell-1)}, j^{(\ell)}\right) \in E(G_c)$ a weight $\omega_{ij}^{(\ell)}$. Then we will analyze the contribution of an edge $(i, j)$ in a single layer $\ell$.
For simplicity, we will assume that $\Phi$ is linear. That is, we take $\varphi^{(\ell)}$ to be the identity in \eqref{eq:gnn-layer}.
Then we may write
\begin{equation}
    \Phi(v;G) = \sum_{u \in \Enn^{(L)}} \partials[z_v]{x_u} \cdot x_u.
\end{equation}
We then decompose the computation graph into paths: writing $\Pi_u^v$ for the set of paths
\begin{equation}
    \pi = \left(u^{(0)} = j_0, j_1, \dots, j_{L-1}, j_L = v^{(L)}\right)
\end{equation}
from $u^{(0)}$ to $v^{(L)}$ in $G_c$, we expand using the chain rule to write
\begin{equation}
    \partials[z_v]{x_u} = \sum_{\pi \in \Pi_u^v} \left[\prod_{\ell = 1}^L \partials[x_{j_\ell}^{(\ell)}]{x_{j_{\ell-1}}^{(\ell-1)}}\right].
\end{equation}
This path decomposition is well-defined because $G_c$ is a DAG.
\begin{remark}
    This path decomposition is employed at the neuron level in \cite{choromanska2015loss, kawaguchi2016deeplearningpoorlocal}. We follow \cite{schnake2021higher} in considering \emph{node-level} walks in the computation graph.
\end{remark}
\begin{proposition}
    \label{prop:edge-contribution}
    For an $L$-layer linear GNN $\Phi$, the contribution of an edge $(i, j)$ in layer $\ell$ is given by
    \begin{equation}
        C^{(\ell)}(i,j) \coloneqq \partials[z_v]{\omega_{i,j}^{(\ell)}} = \partials[z_v]{x_{j}^{(\ell)}} \cdot \partials[x_{j}^{(\ell)}]{x_{i}^{(\ell-1)}} \cdot x_{i}^{(\ell-1)}
        \label{eq:edge-contribution}
    \end{equation}
    in the sense that
    \begin{equation}
        z_v = \sum_{(i, j) \in E^{(\ell)}(G_c(v))} C^{(\ell)}(i,j).
        \label{eq:total-edge-contribution}
    \end{equation}
\end{proposition}
The proof of \cref{prop:edge-contribution}, which we provide in \cref{proof:edge-contribution}, relies on decomposing the output of the network into distinct paths in the computation graph. As we will see in \cref{sec:connections-other}, this path decomposition is central to several methods including GNN-LRP \cite{schnake2021higher} and GOAt \cite{lu2024goat}.

Here the factor $x_i^{(\ell-1)}$ captures the computation graph of $i^{(\ell-1)}$, and the factor $\partials[z_v]{x_{j}^{(\ell)}}$ captures every path in the computation graph after ${j}^{(\ell)}$. In this manner, we can decompose $z_v$ into the contribution of each edge $\left(i^{(\ell-1)}, j^{(\ell)}\right)$ for a given layer $\ell$.
The expression in \eqref{eq:edge-contribution}, which is derived from $\partials[z_v]{\omega_{ij}^{(\ell)}}$, does not rely on the linearity of $\Phi$. Only the second part, \eqref{eq:total-edge-contribution}, uses this assumption.

Notice that in the linear case, \eqref{eq:edge-contribution} does not depend on $\omega_{ij}^{(\ell)}$. Unlike the input-level case where even a linear model can have a nonlinear relationship to the edge weights $\omega_{ij}$, \cref{prop:edge-contribution} shows that $\Phi$ relies linearly on each $\omega_{ij}^{(\ell)}$ when considered separately. This implies the following corollary, which connects edge gradients and Occlusion in the layerwise setting:
\begin{corollary}
    Layerwise edge gradients are equivalent to layerwise edge occlusion for a linear GNN.
    \label{cor:layerwise-grad-occ}
\end{corollary}
The second part of \cref{prop:edge-contribution} also implies the Sensitivity-$n$ property for all $n$ from \cite{ancona2018}. That is, removing any $n$ edges from layer $\ell$ will change $z_v$ by exactly the sum of their attributions. In particular, for $n = |E^{(\ell)}(G_c)|$ this is the \emph{Completeness} property \cite{sundararajan2017axiomatic}.

Analyzing nonlinear networks directly becomes challenging; even ReLU activation becomes difficult to characterize as the activity of each neuron may change in unpredictable ways when an edge is removed. Inactive neurons also lead to vanishing gradients, leaving us unable to determine what effect, if any, an edge might have if it were part of an active path. To deal with this, we borrow notation from GOAt \cite{lu2024goat}, which refers to ``activation patterns''
\begin{equation}
    P^{(\ell)}_{j,k} = \begin{cases}
        \frac{(x_j^{(\ell)})_k}{(z_j^{(\ell)})_k} & (z_j^{(\ell)})_k \neq 0 \\
        0 & \text{otherwise}
    \end{cases}
\end{equation}
where $(z_j^{(\ell)})_k$ is the $k$-th entry of the pre-activated representation of node $j$ at layer $\ell$, and $(x_j^{(\ell)})_k$ refers to the post-activated representation. If $\Phi$ is ReLU-activated, then
\begin{equation}
    \partials[x_{j_\ell}^{(\ell)}]{x_{j_{\ell-1}}^{(\ell-1)}} = {P^{(\ell)}_{j_\ell}}^T W^{(\ell)}.
\end{equation}
We then use the following assumption from \cite{choromanska2015loss, kawaguchi2016deeplearningpoorlocal, xu2018jumping}:
\begin{assumption}
    \label{assump:equal-act-prob}
    Suppose each ReLU unit is activated independently with the same probability $\rho \in (0,1)$. That is, suppose each $P^{(\ell)}_{j,k}$ is a Bernoulli random variable with success probability $\rho$.
\end{assumption}
Under \cref{assump:equal-act-prob},
\begin{equation}
    \E\left[\partials[x_{j_\ell}^{(\ell)}]{x_{j_{\ell-1}}^{(\ell-1)}}\right] = \rho W^{(\ell)}.
\end{equation}
That is, a nonlinear network will behave linearly in expectation.
Although the second part of \cref{prop:edge-contribution} no longer holds when $\Phi$ has nonlinearities, we can justify its utility under the \cref{assump:equal-act-prob} with the following corollary:
\begin{corollary}
    Let $\Phi$ be an $L$-layer sum-aggregated GNN with ReLU activations. Let $\widetilde{\Phi}$ be the network obtained by ignoring each ReLU unit. That is, the linear network with the same weights as $\Phi$. Denote by $\widetilde{C}^{(\ell)}(i,j)$ the contribution of edge $(i, j)$ in layer $\ell$ to $\widetilde{\Phi}$. Then under \cref{assump:equal-act-prob},
    \begin{equation}
        \E[z_v] = \sum_{(i, j) \in E^{(\ell)}(G_c)} \rho^L \widetilde{C}^{(\ell)}(i,j).
        \label{eq:expected-edge-contribution}
    \end{equation}
\end{corollary}

%% file: sections/other-methods.tex
\section{Connections between Other Methods}
\label{sec:connections-other}

Thus far we have seen that perturbation-based explainers can be recast in terms of gradient-based methods under certain assumptions.
Here we point out additional connections between explanation methods. Note that we do not assume linearity in this section.

\subsection{Search-based: EiG-Search}

Here we point out a connection in the input-level setting between the recently proposed EiG-Search method \cite{pmlr-v235-lu24g} and the simple perturbation-based Occlusion method. EiG-Search uses two phases to produce explanations: edges are scored, as in other methods, and then used as a heuristic for a simple linear search to identify an important subgraph. EiG-Search proposes a so-called \emph{linear gradient} score over subgraph components where the importance of an edge set $E_S$ is given by
\begin{equation}
    s(v;E_S) = \frac{\Phi(v;G) - \Phi(v;G_S)}{|A(G) - A(G;E_S)|}
    \label{eq:linear-gradients}
\end{equation}
where
\begin{equation}
    A(G;E_S)_{i,j} = \begin{cases}
        w_{\text{base}} & (i,j) \in E_S \\
        A(G)_{i,j} & (i,j) \in E \setminus E_S \\
        0 & (i,j) \notin E
    \end{cases}
\end{equation}
for a selected baseline weight value $w_{\text{base}}$. In their proposed method, however, $E_S$ is typically taken to be a single edge, and the baseline weight is taken to be zero in line with the zero baseline in \cite{sundararajan2017axiomatic}. In other words, $|A(G) - A(G;E_S)| = 1$, and so we have the following:
\begin{proposition}
    \label{prop:eig-occ}
    For an edge $(i,j)$ in an unweighted graph and baseline weight $w_{\text{base}} = 0$, we have $s(v;\{(i,j)\}) = \operatorname{Occ}_v(i,j)$.
\end{proposition}
\cref{prop:eig-occ} applies to any GNN, since both $s(v;E_S)$ and Occlusion are model-agnostic.

\subsection{Decomposition-based: GNN-LRP}

We turn our attention now to the layerwise setting, where we consider \emph{layerwise relevance propagation} (LRP) for GNNs. The equivalence between gradient backpropagation and LRP is well-known \cite{ancona2018, schnake2021higher}, but the application of LRP to GNNs is made difficult by the number of possible paths in the computation graph being exponential in $L$. Recently, \cite{pmlr-v202-xiong23b} proposes a heuristic search approach to approximate the top node-level walks in $O(|V|^2\bar{p}^2L)$ time, where $\bar{p}$ is the maximum node feature size across all layers. However, we show that one can leverage the observation that the computation graph is a topologically sorted DAG to exactly compute the most relevant walks in comparable time:
\begin{theorem}
    \label{thm:dag-lrp}
    The most relevant node-level walk in a GNN can be computed exactly in $O(L(|V|+|E|)\gamma(\bar{p}))$ time, where $\gamma(\bar{p}))$ is the runtime of any matrix multiplication algorithm.
\end{theorem}
\cref{thm:dag-lrp} applies to any GNN without skip connections \cite{xu2018jumping}.
We provide a proof and a detailed description of the algorithm as \cref{alg:lrp-dag-longest} in \cref{proof:dag-lrp}.
Our \cref{alg:lrp-dag-longest} traverses the computation graph the same manner as AMP-ave-Basic \cite{pmlr-v202-xiong23b}. This also implies a tighter runtime bound of $O(L\bar{p}^2(|V|+|E|))$ for EMP-neu-Basic and AMP-ave-Basic in \cite{pmlr-v202-xiong23b}, which is significantly faster for large sparse graphs where $|V| \ge \bar{p}$.

By pointing out that AMP-ave-Basic uses the longest path algorithm for a topologically sorted DAG, we obtain a simple new interpretation of the algorithm. Furthermore, by leveraging the computation path decomposition, we can compute the \emph{exact} walk scores rather than approximating by averaging the columns of the LRP matrix as in \cite{pmlr-v202-xiong23b}. Essentially, rather than averaging over the neurons at each node as in \cite{pmlr-v202-xiong23b}, we defer the marginalization over neurons until the final layer, where the final layer gradient matrices provide the marginalization weights. 

\subsection{Decomposition-based: GOAt}
Graph Output Attribution (GOAt) \cite{lu2024goat} is an analytical decomposition method that expands the GNN output into the sum of many scalar products. We will consider the GOAt formulation for the GNN given in \eqref{eq:gnn-layer} with ReLU activation, which is of the form
\begin{equation}
    \mu = 
    \left(\prod_{\ell = 1}^L P^{(\ell)}_{\alpha_{\ell,0}, \beta_{\ell,1}}\right)
    \left(\prod_{\ell = 1}^L A^{(\ell)}_{\alpha_{\ell,0}, \alpha_{\ell,1}}\right)
    (x_u)_k \left(\prod_{\ell = 1}^L W^{(\ell)}_{\beta_{\ell,0}, \beta_{\ell,1}}\right)
    \label{eq:goat-expansion}
\end{equation}
where $P^{(\ell)}$ refers to the matrix of activation patterns in layer $\ell$, $A^{(\ell)}$ to the adjacency matrix, $(x_u)_k$ to the $k$-th input feature of node $u$, and $W^{(\ell)}$ to the weight matrix of $\Phi$ in layer $\ell$. The indices $(\alpha_{\ell,0}, \alpha_{\ell,1})$, $(\beta_{\ell,0}, \beta_{\ell,1})$, range over matrix entries in layer $\ell$. Here we omit additional parameters and activations associated with, e.g., additional normalization layers or a final fully connected classifier. We denote an arbitrary component by $\nu$. We will denote by $\# \mu$ the number of \emph{distinct} factors in $\mu$ (for example, if $\mu = x^2y$, then $\# \mu = 2$). Then each component $\nu$ is given the attribution
\begin{equation}
    I_\nu(\Phi) = \sum_{\mu \ni \nu} \frac{\prod_{\nu' \in \mu} \nu}{\# \mu}.
    \label{eq:goat-attribution}
\end{equation}
Lu et al.~\cite{lu2024goat} further adjust \cref{eq:goat-attribution} to account for the fact that activation patterns $P^{(\ell)}$ also depend on other components, and additionally isolate the contribution from bias terms in the network.

Again the decomposition of the computation graph into paths illustrates the connections between this method and another. Notice that each scalar product in \eqref{eq:goat-attribution} is exactly a \emph{neuron-level walk} as described in \cite{schnake2021higher}. Thus the summands can be thought of as GNN-GI scores from \cite{schnake2021higher}, which are given by
\begin{equation}
    R_\mu^{\text{GNN-GI}} \coloneqq \left(\prod_{\ell=1}^L \partials[(x_{j_\ell})_{k_\ell}]{(x_{\ell-1})_{k_{\ell-1}}}\right) \cdot (x_u)_{k_0}
    \label{eq:gnn-gi}
\end{equation}
for the neuron-level walk $((x_{j_\ell})_{k_\ell})_{\ell=1}^L$ corresponding to $\mu$. Then:
\begin{proposition}
    \label{prop:goat-lrp}
    For a component $\nu$, we have
    \begin{equation}
        I_\nu(\Phi) = \sum_{\mu \ni \nu} \frac{R_\mu^{\text{GNN-GI}}}{\# \mu}.
    \end{equation}
\end{proposition}
We will omit the rest of the analysis from \cite{lu2024goat}, which adjusts for the relevance attributed to bias terms and the relevance of $\nu$ to activation patterns $P^{(\ell)}$. While GOAt typically requires expert knowledge of model parameters and careful computation of scalar product attributions, this connection between GOAt and GNN-GI (and hence with LRP) implies that GOAt can be computed using algorithms developed for LRP such as \cite{pmlr-v202-xiong23b} and our \cref{alg:lrp-dag-longest}.

%% file: sections/experiments.tex
\section{Experiments}
\label{sec:experiments}

To validate our theoretical insights, we compare the behavior of different GNN explainability methods across multiple experiments.  We introduce a very simple baseline called \emph{Positive Gradients}, which will give an edge an attribution of 1 if $\partials[Y_v]{\omega_{ij}} > \varepsilon$ and 0 otherwise. We use $\varepsilon = 0.001$ in Negative Evidence and $\varepsilon = 0$ elsewhere. (We do not see any appreciable effect in any experiment other than Negative Evidence, and we will explain this difference in behavior in the following subsection.) We also use a Random Edges baseline, which assigns a random attribution in $[0,1]$ to each edge. Since GNNExplainer only provides input-level explanations, we also implement a Layerwise GNNExplainer which jointly trains a mask for each layer, similar to the unamortized version of GraphMask \cite{schlichtkrull2022interpretinggraphneuralnetworks}.

Through our experiments, we show that GNNExplainer can be approximated by the much simpler Positive Gradients on simple models, as suggested by \cref{prop:no-grad-flip}. We also show that Layerwise Gradients approximate Layerwise Occlusion, as predicted by \cref{cor:layerwise-grad-occ}.
We quantify the similarity between explanations using the cosine similarity of their edge masks and report the average and standard deviation over each example in the test set. We adopt cosine similarity to avoid the effects of differing sparsity for different instances in the same dataset and to avoid rescaling methods whose edge masks are not normalized between 0 and 1.

\subsection{Synthetic Data}

\subsubsection{Negative Evidence}
\label{subsec:neg-evidence}

We use the negative evidence experiment from \cite{faber2021}. We generate 4 training graphs and one testing graph. Each graph has 2000 nodes. There are 10 red nodes and 10 blue nodes, with the rest of the nodes colored gray. A 1-layer linear sum-aggregated GNN with no bias terms is trained to predict if a node has more red or blue neighbors, achieving perfect accuracy.

\begin{figure}
    \centering
    \includegraphics[width=\linewidth]{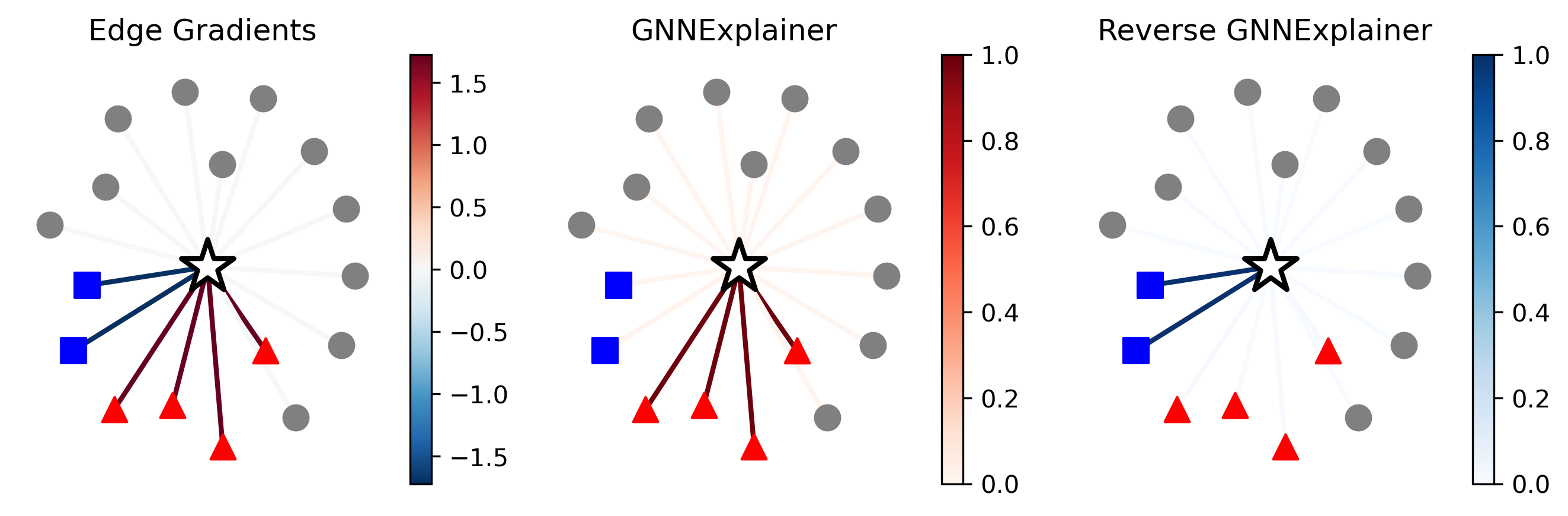}
    \caption{Explainer predictions on Negative Evidence using Edge Gradients (left), GNNExplainer (middle), and GNNExplainer with the target class reversed (right). Edge Gradients recover the ground truth.}
    \label{fig:neg-evidence}
    \Description{A star graph where the target node has four red neighbors, two blue neighbors, and several gray neighbors.}
\end{figure}

\begin{figure}
    \centering
    \includegraphics[width=.66\linewidth]{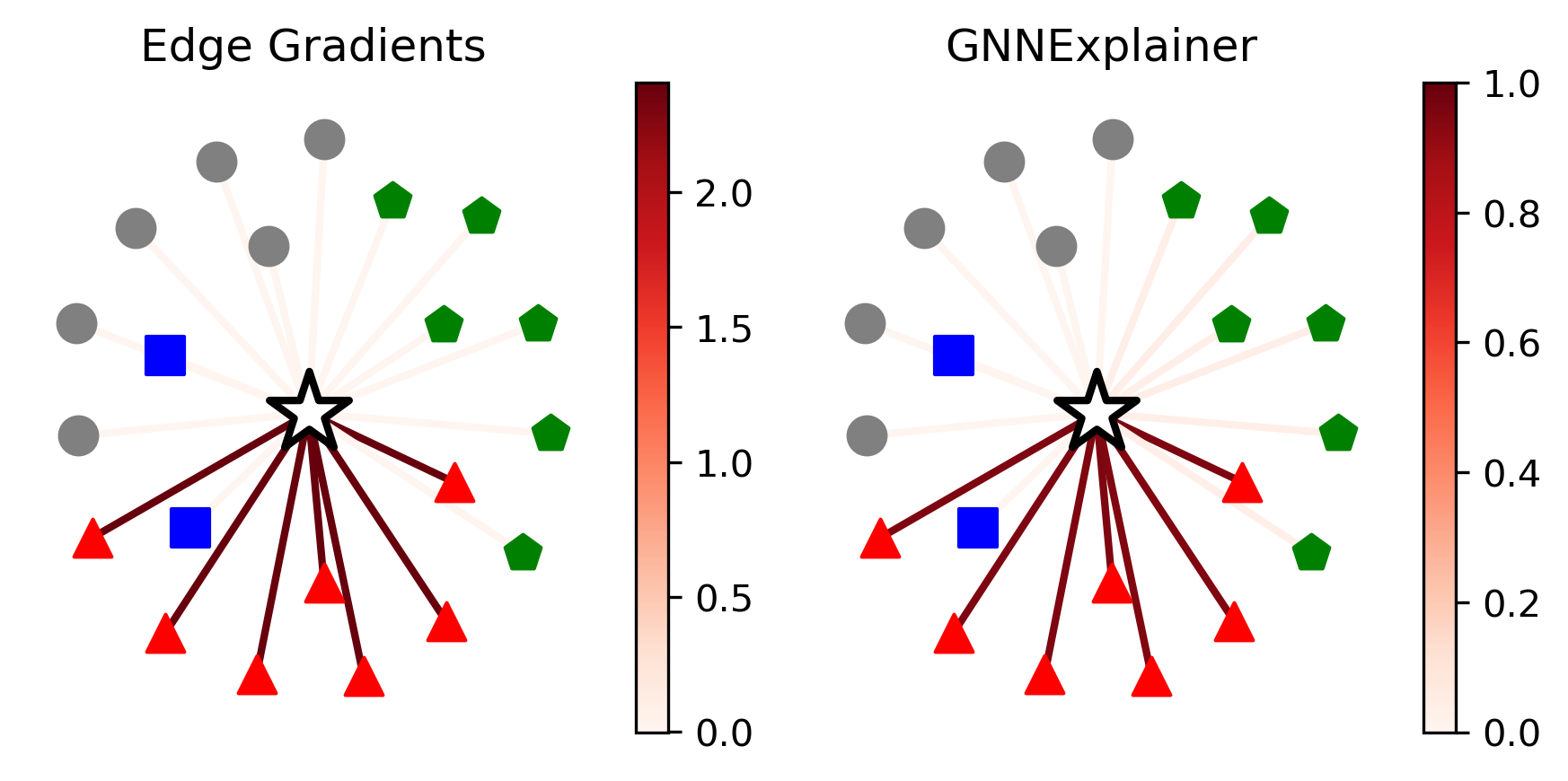}
    \caption{Explanations on our multiclass version of Negative Evidence using Edge Gradients (left) and GNNExplainer (right). Notice that in this case, Edge Gradients does not assign negative attribution to non-red edges.}
    \label{fig:neg-evidence-multiclass}
    \Description{A star graph where the target node has seven red neighbors, two blue neighbors, six green neighbors, and several gray neighbors.}
\end{figure}

\paragraph{GNNExplainer keeps positive evidence.}
As predicted by \cref{prop:1-layer}, GNNExplainer selects exactly the edges whose gradients are positive (\cref{fig:neg-evidence}). Note that GNNExplainer and Edge Gradients differ due to negative evidence: GNNExplainer gives negative edges an attribution of zero as in \cref{prop:1-layer}. On the other hand, irrelevant edges have gradients very close to zero, and thus their mask values should stay near their initialization. However, GNNExplainer's regularization term $\alpha \norm{\Omega}_1$ creates an additional negative gradient in its optimization, forcing irrelevant edges towards zero. By setting $\beta = 0$ and $\varepsilon = \alpha = 0.001$, we threshold away irrelevant edges as well as negative edges. Using this simple gradient heuristic, we recover the GNNExplainer masks almost exactly: across each test node in the graph, the average cosine similarity between edges with positive gradient and GNNExplainer is $0.9979 \pm 0.0029$.

\paragraph{Negative Evidence with GNNExplainer.}
Although GNNExplainer's inability to distinguish negative evidence from irrelevant evidence is potentially undesirable \cite{faber2021}, \cref{prop:no-grad-flip} implies a remedy: by changing the target class for binary classification, GNNExplainer will remove the positive evidence for the original class and instead keep the negative evidence. \cref{fig:neg-evidence} shows an example where GNNExplainer matches the positive gradients and reversing the target class for GNNExplainer matches the negative gradients. In fact, subtracting the Reversed GNNExplainer explanations from the original GNNExplainer explanations gives an average of $0.9741\pm 0.0315$ cosine similarity with the raw edge gradients.

We also perform a similar experiment using a multiclass classification task where nodes may be red, blue, or green (\cref{fig:neg-evidence-multiclass}). Across each test node in the graph, the average cosine similarity between edges with positive gradient and GNNExplainer is $0.9977 \pm 0.0038\%$. Notice, however, that the edge gradients no longer highlight edges to other colors as having negative contribution, as in this setting the model can learn positive evidence for each class separately from negative evidence for the others.

\subsubsection{Infection}
\label{subsec:infection}

\begin{figure}
    \centering
    \includegraphics[width=.6\linewidth]{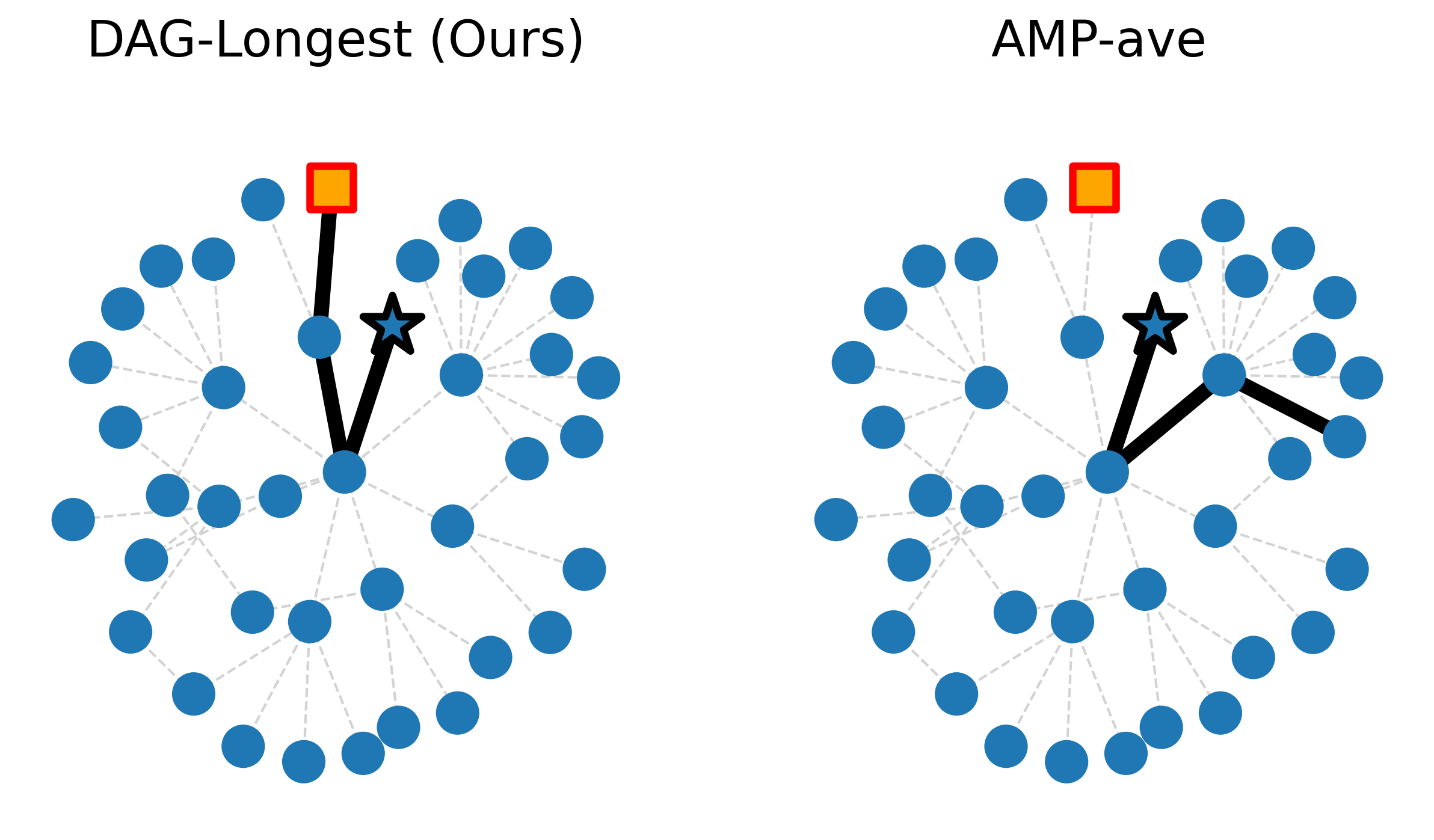}
    \caption{Best path produced by \cref{alg:lrp-dag-longest} (left) and AMP-ave-Basic (right) to explain the infection distance for the star node. \cref{alg:lrp-dag-longest} recovers the ground truth, identifying 3 hops to the infected square node.}
    \label{fig:infection-example}
    \Description{Two Infection graphs with identified paths highlighted.}
\end{figure}

\begin{figure}
    \centering
    \includegraphics[width=.38\linewidth]{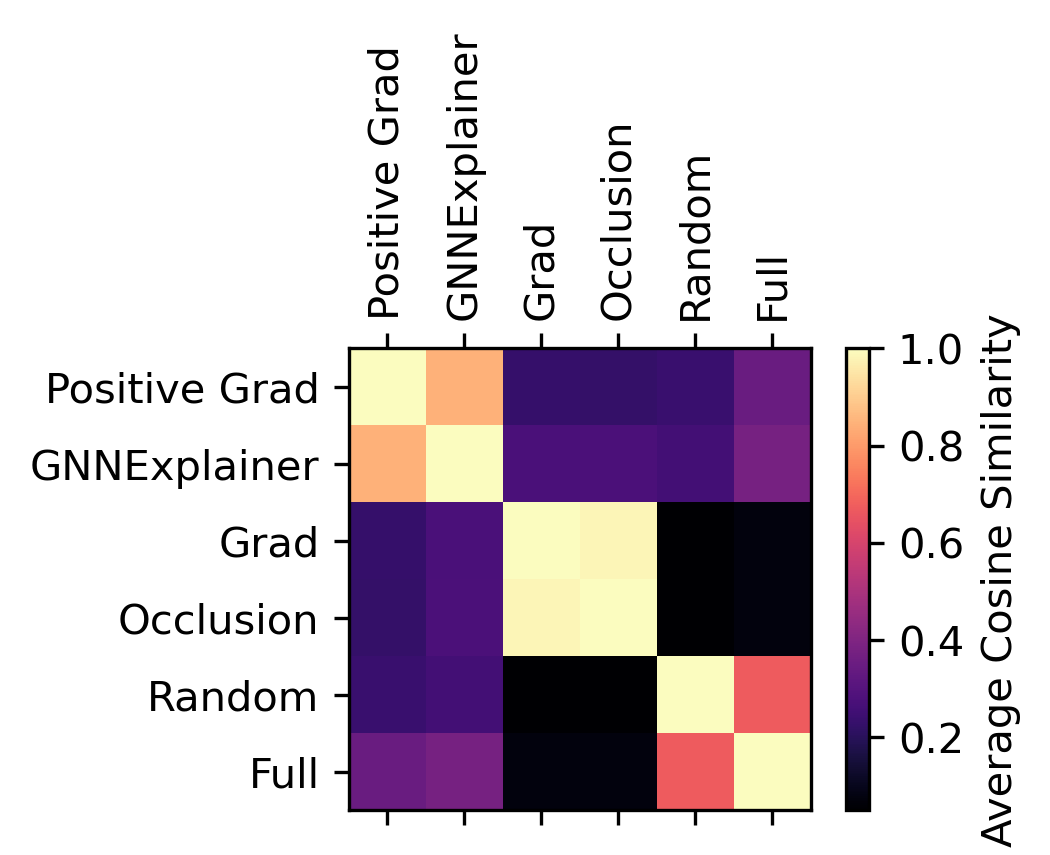}
    \caption{Pairwise average cosine similarity between input-level explanations on Infection.}
    \label{fig:infection-sims}
    \Description{A matrix of pairwise cosine similarity values. Most values are low except Positive Gradients with GNNExplainer and Edge Gradients with Occlusion.}
\end{figure}

\begin{figure}
    \centering
    \includegraphics[width=\linewidth]{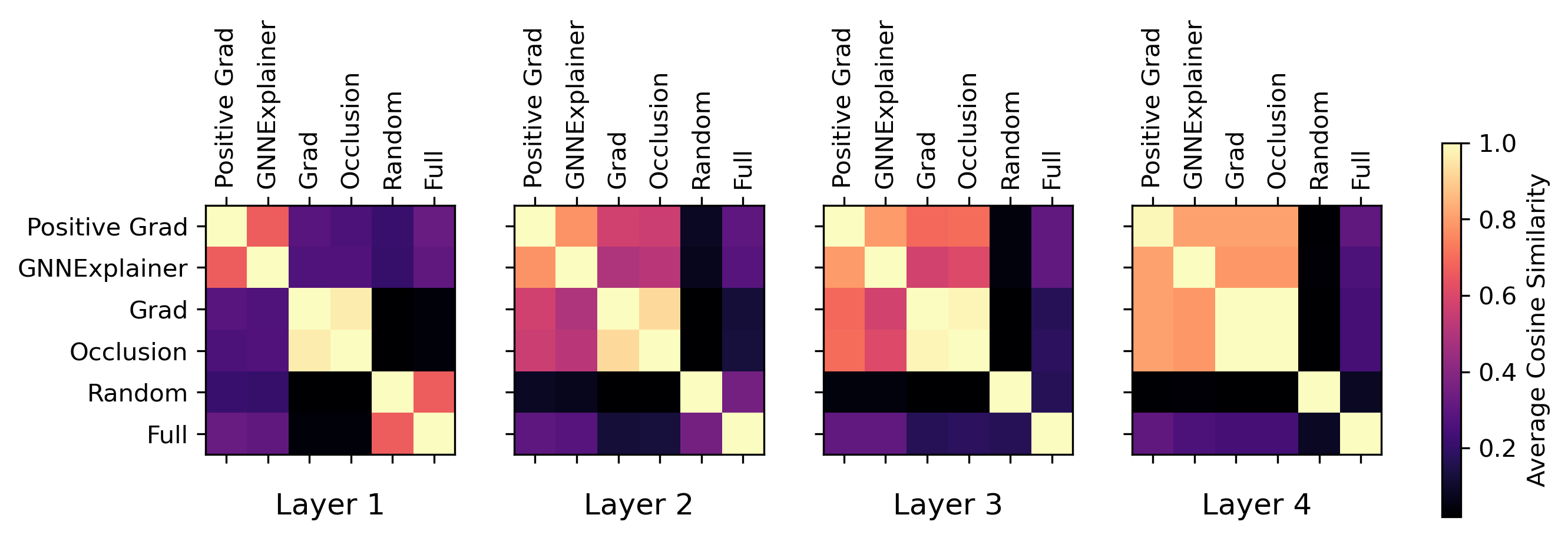}
    \caption{Pairwise average cosine similarity between layerwise explanations on Infection on each layer.}
    \label{fig:infection-layerwise-sims}
    \Description{Four matrices of pairwise cosine similarity values. Edge Gradients and Occlusion consistently have the highest similarities, followed by Positive Gradients and GNNExplainer. Layer 4 also has high similarity between all four of Positive Gradients, GNNExplainer, Edge Gradients, and Occlusion.}
\end{figure}

In this task we train a 4-layer sum-aggregated GNN on the version of the infection dataset from \cite{baldassarre2019explainabilitytechniquesgraphconvolutional, faber2021}. We generate random Erdős-Renyi graphs ($p = 0.004$) of 1000 nodes, each containing 50 infected nodes and 950 healthy nodes. The task is to predict the length of the shortest directed path from an infected node to the target node, with 6 classes: 0 (the node is already infected) to 5 (the node is $\geq 5$ hops from an infected node). We generate 4 training graphs and one graph for testing and explanation. The model achieves perfect accuracy on the test graph.

\paragraph{Exact DAG path search outperforms AMP-ave.}
Because the ground truth is in the form of paths, we evaluate our DAG-longest-path algorithm (\cref{alg:lrp-dag-longest}). Following \cite{faber2021}, we explain only on nodes whose class is not $5+$, as they have no ground truth explanations. We also restrict our explanations to nodes with a unique ground truth infection path to avoid ambiguity. Comparing our \cref{alg:lrp-dag-longest} (described in \cref{proof:dag-lrp}) to AMP-ave-Basic \cite{pmlr-v202-xiong23b}, we demonstrate the benefit of our exact computation: \cref{alg:lrp-dag-longest} successfully recovers $99.07\%$ of the exact ground truth paths, compared to $91.12\%$ for AMP-ave-Basic without significant difference in runtime ($8.12 \pm 5.48$s for our method vs. $8.18 \pm 5.52$s for AMP-ave).
(For our comparison, we use gradients as specified in \cref{alg:lrp-dag-longest}, which is equivalent to $0$-LRP \cite{ancona2018, schnake2021higher}. Note that we do not assume the gradients are precomputed as in \cite{pmlr-v202-xiong23b}.) We provide an example in \cref{fig:infection-example} where AMP-ave takes a ``wrong turn'' due to the approximation error in layer 3. In contrast, our exact DAG-longest path search produces the ground truth.

\paragraph{Comparing Explanations}
We compare edge masks produced by six explanation methods on the test graph: Positive Gradients, GNNExplainer, Edge Gradients, Occlusion, Random Edges, and a full edge mask for the 4-hop neighborhood. \cref{fig:infection-sims} shows the pairwise cosine similarity averaged over each example in the test set. GNNExplainer shows a high similarity to positive gradients, at $0.8472 \pm .0672$. We also examine the layerwise attributions in \cref{fig:infection-layerwise-sims}. Again we see strong similarity between Positive Gradients and GNNExplainer, demonstrating the implications of \cref{prop:no-grad-flip}. We also see that Layerwise Grad and Layerwise Occlusion have very high similarity across all layers, validating \cref{cor:layerwise-grad-occ}.

\subsection{Real Data}
\label{subsec:real-data}

We train three popular architectures of GNN: GCN \cite{kipf2017semisupervisedclassificationgraphconvolutional}, GAT \cite{velivckovic2017graph, brody2021attentive}, and GIN \cite{xu2018powerful}. For each architecture, we train a ReLU-activated network of $L = 1, \dots, 5$ layers on each of five node classification datasets and three graph classification datasets. For node classification we use the citation networks Cora and Citeseer from \cite{yang2016revisiting} and the website networks Cornell, Texas, and Wisconsin from \cite{pei2020geom}. For graph classification we use the molecular datasets MUTAG and PROTEINS from \cite{morris2020tudatasetcollectionbenchmarkdatasets}, as well as Graph-SST2 from \cite{yuan2022explainability} We provide additional details in \cref{appdx:training}.

\begin{figure}
    \centering
    \includegraphics[width=.6\linewidth]{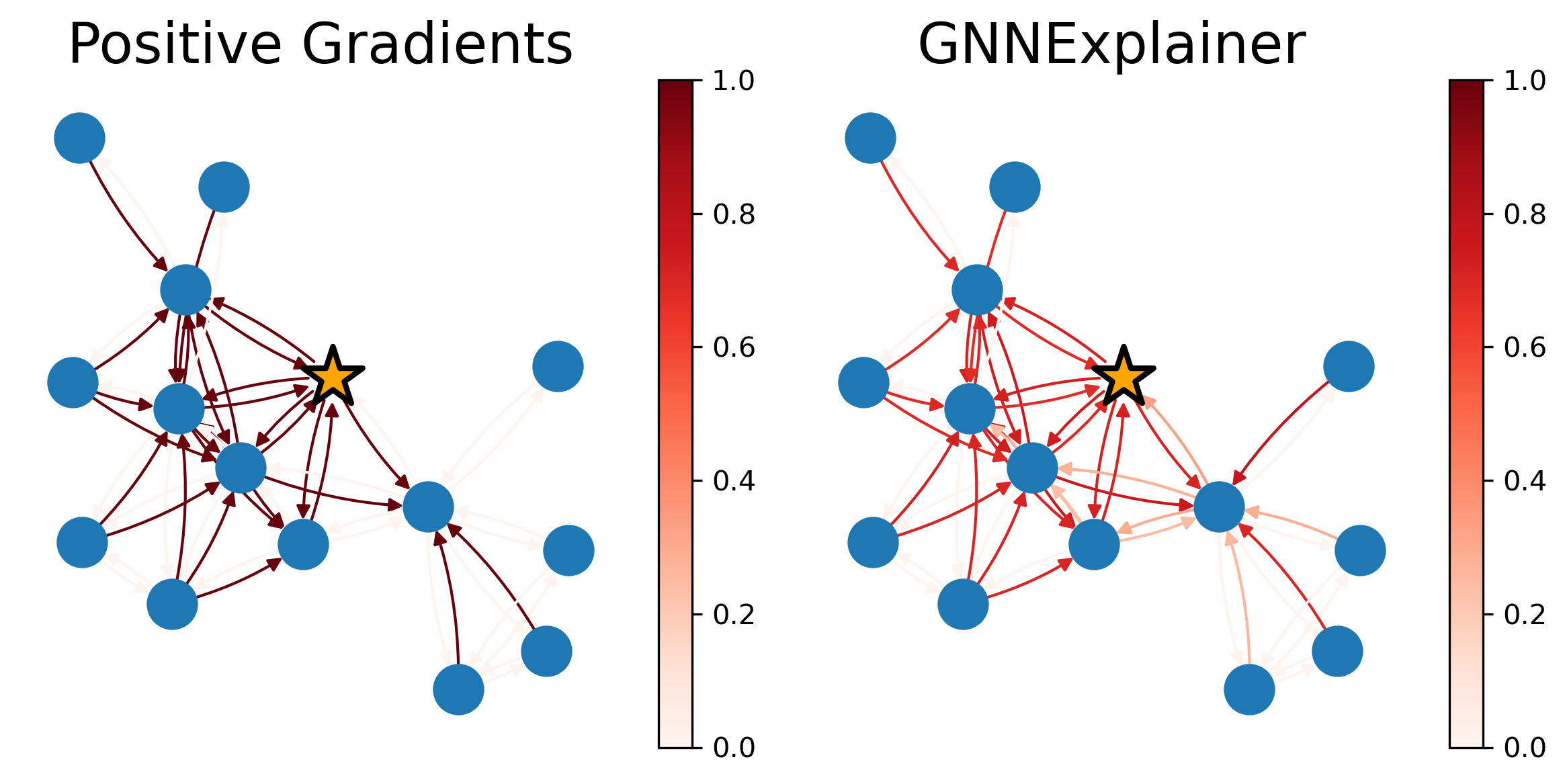}
    \caption{Positive Gradients (left) and GNNExplainer explanation (right) for the prediction of a 2-layer GCN on a test node in Cora. The explanations have a similarity of 0.9570.}
    \Description{Two graphs, each with explanation results highlighted in red. Both graphs have similar edges highlighted.}
    \label{fig:cora-qualitative}
\end{figure}

\begin{figure*}[ht]
    \centering
    \includegraphics[width=\linewidth]{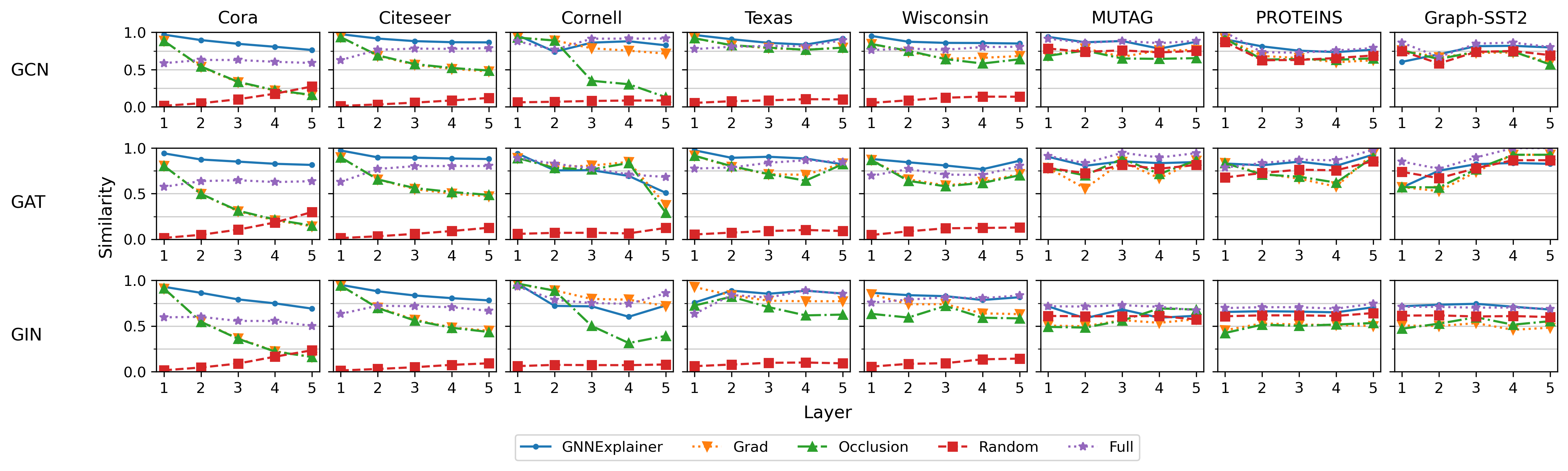}
    \caption{Average input-level similarity between explanation methods (GNNExplainer, Edge Gradients, Occlusion, Random, and Full) and the Positive Gradients baseline. Averages are taken over each test example.}
    \label{fig:real-pos-grad}
    \Description{24 line graphs, one for each combination of dataset and architecture. Average cosine similarity to Positive Gradients is plotted against the depth of the target network. 5 lines are shown. GNNExplainer consistently has high similarity, usually starting near 1.0 at 1 layer and slowly falling.}
\end{figure*}

\begin{figure*}
    \centering
    \includegraphics[width=\linewidth]{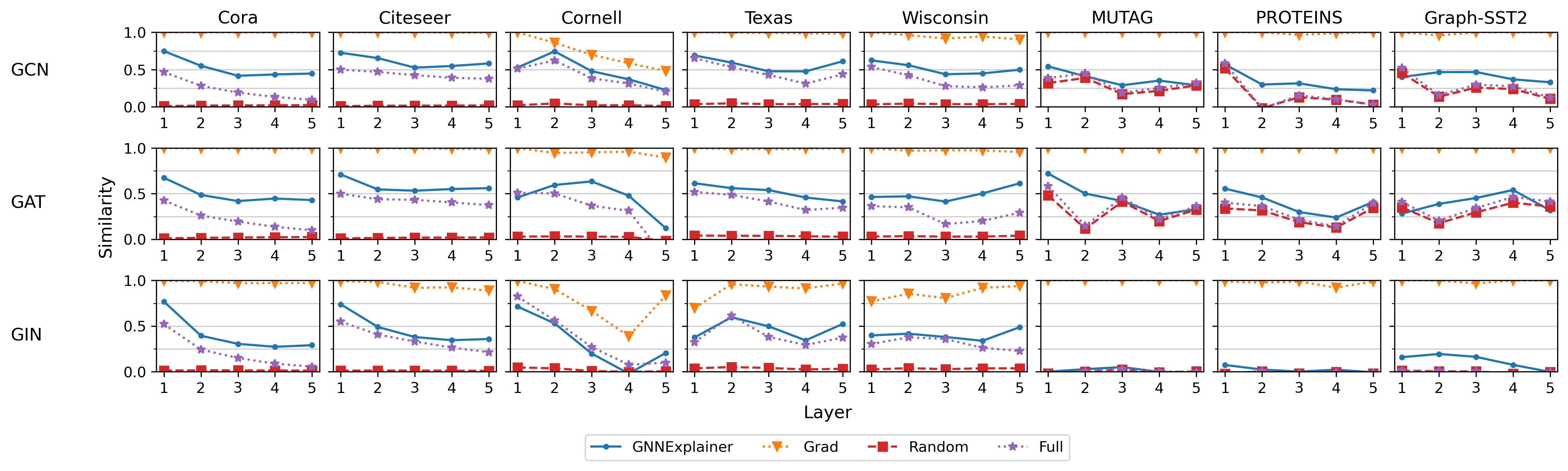}
    \caption{Average layerwise similarity between explanation methods (GNNExplainer, Edge Gradients, Random, and Full) and Occlusion. Averages are taken over each test example and each layer.}
    \label{fig:real-layerwise-occlusion}
    \Description{24 line graphs, one for each combination of dataset and architecture. Average cosine similarity to Occlusion is plotted against the depth of the target network. 4 lines are shown. Edge Gradients consistently have high similarity, usually remaining near constant at 1.0.}
\end{figure*}

\paragraph{Comparing Explanations.}
We show how \cref{prop:no-grad-flip} creates GNNExplainer attributions which are similar to Positive Gradients. For a qualitative example, we compare the GNNExplainer explanation for a node in Cora with the Positive Gradients explanation on the same instance (\cref{fig:cora-qualitative}). Quantitatively, \cref{fig:real-pos-grad} shows the average cosine similarity of explanations to Positive Gradients on each GNN across all test examples. This indicates that GNNExplainer displays an insensitivity to the size of the gradients, which may account for Edge Gradients outperforming GNNExplainer on some real datasets (e.g. \cite{pmlr-v198-amara22a}). We also validate \cref{cor:layerwise-grad-occ} across architectures on each dataset in \cref{fig:real-layerwise-occlusion}, showing the similarity of each method to Layerwise Occlusion.

\begin{figure*}
    \centering
    \includegraphics[width=\linewidth]{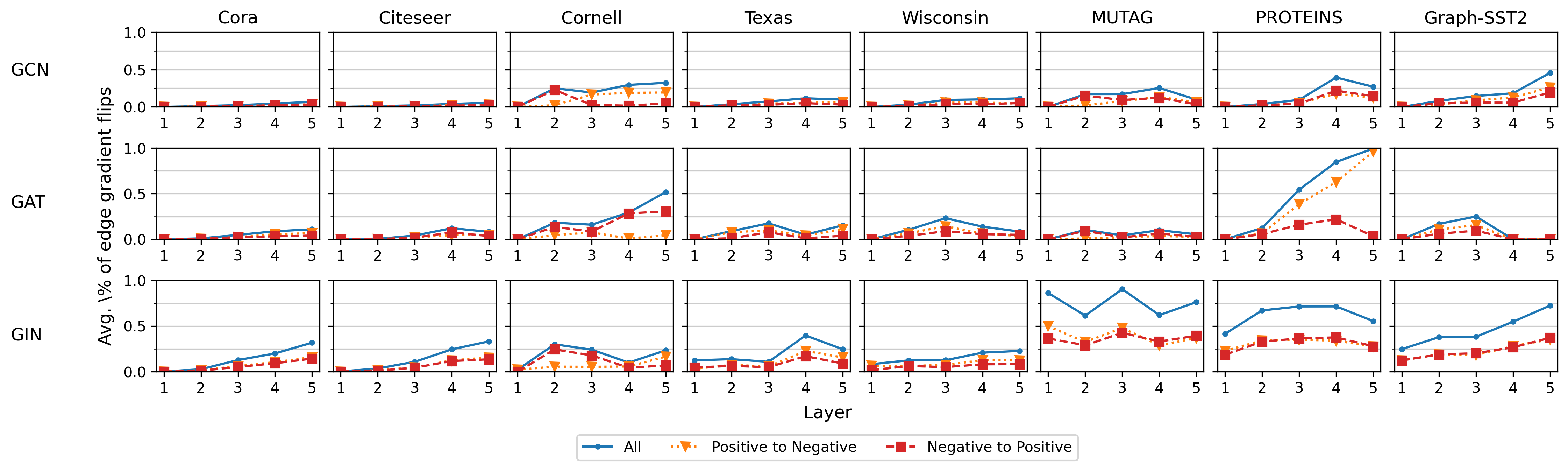}
    \caption{Percent of edges whose gradient changes sign as edge weight varies from 1 to 0. We plot the percent of edges that show any flip, percentage of gradients that flip from positive to negative, and gradients that flip from negative to positive.}
    \label{fig:grad-flips}
    \Description{24 line graphs, one for each combination of dataset and architecture. The average percent of sign flips is plotted against the depth of the target network. 3 lines are shown. For node classification, lines are close to zero. For graph classification, the total flips increases more quickly in some plots. GAT-Proteins reaches 100\% flip rate, with almost all being positive-to-negative. GIN plots reach >70\% flips, split mostly evenly between positive-to-negative and negative-to-positive.}
\end{figure*}

\paragraph{Gradients and Sign-Flipping}
We examine how often the assumption in \cref{prop:no-grad-flip} holds. We vary the mask values of each edge and compute the edge gradients for each test prediction and plot in \cref{fig:grad-flips} how many edge gradients change sign across different values of $\omega_{ij} \in \{1.0, 0.95, \cdots, 0.05, 0.0\}$. For node classification, only a small percentage of edges flip, with slight increases as networks become more nonlinear. For graph classification, almost all edge gradients flip between $\omega_{ij} = 0$ and $\omega_{ij} = 1$. However, for most networks there are more flips from positive to negative than from negative to positive. Since GNNExplainer's default behavior is to promote sparse, binary edge masks with its regularization, GNNExplainer may still discard edges whose gradients flip from negative at $\omega_{ij} = 1$ to positive and keep edges with positive gradients at $\omega_{ij} = 1$, explaining the similarity we observe in \cref{fig:real-pos-grad}.

%% file: sections/conclusion.tex
\section{Conclusion}
\label{sec:conclusion}

\balance
We analyze GNN explainability methods from a theoretical perspective, identifying similarities in both the input-level and layerwise settings. At the input level, our analysis reveals situations in which GNNExplainer approximates our simple Positive Gradients heuristic. In the layerwise setting, we point out that layerwise edge gradients are equivalent to layerwise occlusion in linear networks. We also point out connections between other methods, including a simple reinterpretation of AMP-ave that provides an exact algorithm with improved performance.

We note that our theoretical analysis has limitations: we mostly consider linear GNNs performing binary classification. We also consider edges individually, potentially failing to capture more intricate dependencies between edges during joint optimization. Despite these limitations, we demonstrate both of these phenomena empirically across a variety of datasets, model depths, and architectures. Future work could extend our theoretical results to more general architectures and settings, as well as analyze the behavior of other graph explanation paradigms, such as parameterized perturbation-based methods.
We hope this theoretical perspective will motivate further explainability research for GNNs to leverage the ease of gradient computation, and to provide a more unified perspective on GNN explainability.

%% file: appendix/proofs.tex
\section{Proofs}
\label{appdx:proofs}

\subsection{Proof of Proposition \ref{prop:1-layer}}
\label{proof:1-layer}
\begin{proof}
    Following GNNExplainer, we interpret the output $Y_v|_{G_S}$ as the expectation of a Bernoulli random variable. We write the mutual information in terms of the entropy
    \begin{equation}
        I(Y_v; G_S) = H(Y_v) - H(Y_v \mid G_S)
        \label{eq:mutual-information}
    \end{equation}
    where $H(Y_v)$ is the entropy of the (fixed) GNN prediction and $H(Y_v \mid G_S)$ is the conditional entropy of the soft mask relaxation of $G_S$. That is, each edge $(i,j)$ is assigned a Bernoulli random variable with parameter $\omega_{ij}$ and $G_S$ is sampled from the distribution over the edges defined by $\omega_{ij}$. Then we consider the problem in expectation:
    \begin{equation}
        H(Y_v \mid G_S) = \E\left[
            -\log\sigma\left(
                \sum_{u \in \Enn(v)} \partials[z_v]{x_u} \cdot x_u \cdot \1\{(u,v) \in G_S\}
            \right)
        \right].
        \label{eq:one-layer-entropy}
    \end{equation}
    We see that $H(Y_v \mid G_S)$ is minimized when the sum $\sum_{u \in G_S} \partials[z_v]{x_u} \cdot x_u$ is maximized. This occurs precisely when $G_S$ contains exactly the nodes $u$ and edges $(u,v)$ for which $\partials[z_v]{x_u} \cdot x_u > 0$.
\end{proof}

\begin{remark}
    \citet{ying2019gnnexplainer} further apply Jensen's inequality and minimize $H(Y \mid \E[G_S])$
    where $\E[G_S]$ is specified by the trained soft mask $\Omega$. Despite the potential nonconvexity of $\Phi$, the authors note that the objective often produces successful explanations. In the one-layer case of \cref{eq:one-layer-entropy}, this objective becomes
    \begin{equation}
        \min_{\Omega} H(Y \mid \E[G_S]) = \E\left[
            -\log\sigma\left(
                \sum_{u \in \Enn(v)} \partials[z_v]{x_u} \cdot \omega_{uv} x_u
            \right)
        \right]
    \end{equation}
    from which we ecover \cref{eq:one-layer-entropy} by $\omega_{uv} = \1\{ \partials[z_v]{x_u} \cdot x_u > 0 \}$.
\end{remark}

\subsection{Proof of Proposition \ref{prop:no-grad-flip}}
\label{proof:no-grad-flip}
\begin{proof}
    Recall that we consider binary classification and that without loss of generality, we may take $Y_v = +$. We begin by writing
    \begin{align}
        I(Y_v \mid \omega_{ij} = \omega_{ij}^*) &= H(Y_v) - H(Y_v \mid \omega_{ij} = \omega_{ij}^*) \\
        &= H(Y_v) - \E\left[
            -\log P_\Phi(Y_v \mid \omega_{ij}^*)
        \right] \\
        &= H(Y_v) + \E\left[
            \log \sigma(z_v|_{\omega_{ij}^*}).
        \right]
    \end{align}
    Thus if $\partials[z_v]{\omega_{ij}} > 0$ for all $\omega_{ij} \in [0,1]$, $z_v$ is an increasing function of $\omega_{ij}$, and hence so is the mutual information. Similarly, if $\partials[z_v]{\omega_{ij}} < 0$ for all $\omega_{ij} \in [0,1]$, $z_v$ is a decreasing function of $\omega_{ij}$, and hence so is the mutual information. This completes the proof.
\end{proof}

\subsection{Proof of Proposition \ref{prop:edge-contribution}}
\label{proof:edge-contribution}
\begin{proof}
We use the technique from \cite{xu2018jumping} of decomposing the final output $z_v$ into each path $\pi$ in the set $\Pi_u^v$ of length-$L$ walks in $G$ which begin at a source node $u$ and end at $v$. Equivalently, $\Pi_u^v$ can be thought of as the set of maximal paths between nodes $u^{(0)}$ and $v^{(L)}$ in the computation graph $G_c$. Then we write
\begin{equation}
    z_v =
        \sum_{u \in \Enn^{(L)}(v)} \left[\sum_{\pi \in \Pi_u^v} \prod_{\ell = 1}^L \omega_{j_{\ell-1}, j_\ell}^{(\ell)}\partials[x_{j_\ell}^{(\ell)}]{x_{j_{\ell-1}}^{(\ell-1)}}\right] \cdot x_u.
    \label{eq:path-decomposition}
\end{equation}
    From the summation we can see that the predicted class logit $z_v$ can be decomposed into the contributions from each length-$L$ walk between each input node and $v$. Each path $\pi = (j_\ell)_{\ell=0}^L$ contributes to the prediction
\begin{equation}
   \prod_{\ell = 1}^L \left[\omega_{j_{\ell-1}, j_\ell}^{(\ell)}\partials[x_{j_\ell}^{(\ell)}]{x_{j_{\ell-1}}^{(\ell-1)}}\right] \cdot x_{j_0}^{(0)}.
\end{equation}
Thus we can consider the individual contribution of each edge by considering the contribution of each path which flows through that edge.
Notice that in a linear network,
\begin{align}
    \partials[z_v]{\omega_{i, j}^{(\ell)}}
    &= \sum_{\pi \ni ({i}^{(\ell-1)}, {j}^{(\ell)})}
    \left[
        \prod_{k = \ell+1}^{L} \left(\omega_{j_{k-1}, j_k}^{(k)} \partials[x_{j_k}^{(k)}]{x_{j_{k-1}}^{(k-1)}}\right)
        \cdot \partials[x_{j}^{(\ell)}]{x_{i}^{(\ell-1)}}\right. \\
        &\left.\hspace{2.5cm}\cdot \prod_{k = 1}^{\ell-1} \left(\omega_{j_{k-1}, j_k}^{(k)} \partials[x_{j_k}^{(k)}]{x_{j_{k-1}}^{(k-1)}}\right)
        \cdot x_{j_0}^{(0)}\right] \\
    &= \left[
        \sum_{\pi \in \Pi_{j}^v} \prod_{k = \ell+1}^{L} \omega_{j_{k-1}, j_k}^{(k)} \partials[x_{j_k}^{(k)}]{x_{j_{k-1}}^{(k-1)}}
    \right]
    \cdot \partials[x_{j}^{(\ell)}]{x_{i}^{(\ell-1)}} \\
    &\phantom{==}\cdot \left[
        \sum_{u \in \Enn^{(\ell-1)}(j)}\sum_{\pi \in \Pi_{j}^v} \prod_{k = 1}^{\ell} \omega_{j_{k-1}, j_k}^{(k)} \partials[x_{j_k}^{(k)}]{x_{j_{k-1}}^{(k-1)}} \cdot x_u
    \right] \\
    \intertext{which, evaluated at $\omega_{j_{\ell-1},j_\ell}^{(\ell)} = 1$ becomes}
    &= \partials[z_v]{x_{j}^{(\ell)}} \cdot \partials[x_{j}^{(\ell)}]{x_{i}^{(\ell-1)}} \cdot x_{i}^{(\ell-1)}
\end{align}
Moreover, restricting our attention to layer $(\ell-1)$, we see that
\begin{equation}
    z_v = \sum_{u^{(\ell-1)} \in G_c^{(\ell-1)}} \partials[z_v]{x_{j}^{(\ell)}} \cdot \partials[x_{j}^{(\ell)}]{x_{i}^{(\ell-1)}} \cdot x_{i}^{(\ell-1)}.
\end{equation}
\end{proof}

\subsection{Proof of Theorem \ref{thm:dag-lrp}}
\label{proof:dag-lrp}

\begin{proof}
    \cref{alg:lrp-dag-longest} relies on the classic algorithm for the optimal path in a topologically-sorted DAG. The score for a walk $\pi = (j_0, j_1, \dots, j_{L-1}, j_L)$
    in $G_c$ is
    \begin{equation}
        \left[\prod_{\ell = 1}^L \partials[x_{j_\ell}^{(\ell)}]{x_{j_{\ell-1}}^{(\ell-1)}}\right] \cdot x_{j_0}
    \end{equation}
    which is exactly the GNN-GI attribution in \cite{schnake2021higher}. By replacing the gradient matrices $\partials[x_{u}^{(\ell)}]{x_{u'}^{(\ell-1)}}$ with LRP matrices, one can obtain the exact most relevant walk by LRP score.
    
    Since $G_c$ has $O(L|V|)$ nodes and $O(L|E|)$ edges, \cref{alg:lrp-dag-longest} performs $O(L(|V|+|E|)$ matrix multiplications. Thus the runtime is $O(L(|V|+|E|)\gamma(\bar{p}))$, where $\gamma(\bar{p}))$ is the runtime of any matrix multiplication algorithm.
\end{proof}

\algsetup{
linenosize=\footnotesize,
linenodelimiter=.
}
\begin{algorithm}
\caption{DAG-longest path for top relevant walk}\label{alg:lrp-dag-longest}
    \begin{algorithmic}[1]
        \REQUIRE $G_c$, $\partials[x_{u}^{(\ell)}]{x_{u'}^{(\ell-1)}}$, $x_u^{(\ell)}$ for $\ell = 1, \dots, L$
        \STATE $\parent(u^{(\ell)}) \gets \texttt{None}$ for each $u^{(\ell)}$
        \STATE $\gradcum(u^{(\ell)}) \gets \texttt{None}$ for each $u^{(\ell)}, \ell = 0, \dots, L-1$
        \STATE $\score(u^{(\ell)}) \gets -\infty$ for each $u^{(\ell)}, \ell = 0, \dots, L-1$
        \STATE $\gradcum(u^{(\ell)}), \score(u^{(L)}) \gets \partials[z]{x_u^{(L)}}$ for each $u^{(L)}$
        \FOR{$\ell = L, \dots, 1$}
            \FOR{$u^{(\ell)} \in \Enn^{(\ell)}(v)$}
                \FOR{$({u'}^{(\ell-1)}, u^{(\ell)}) \in E^{(\ell)}(G_c)$}
                \STATE new\_score $\gets \gradcum(u^{(\ell)}) \partials[x_{u}^{(\ell)}]{x_{u'}^{(\ell-1)}} x_u^{(\ell-1)}$
                    \IF{new\_score $< \score({u'}^{(\ell-1)})$}
                        \STATE $\parent({u'}^{(\ell-1)}) \gets u^{(\ell)}$
                        \STATE $\score({u'}^{(\ell-1)}) \gets$ new\_score
                        \STATE $\gradcum({u'}^{(\ell-1)}) \gets \gradcum(u^{(\ell)}) \partials[x_{u}^{(\ell)}]{x_{u'}^{(\ell-1)}}$
                    \ENDIF
                \ENDFOR
            \ENDFOR
        \ENDFOR
        \STATE ${u^*}^{(0)} \gets \argmax_{u^{(0)}} \score(u^{(0)})$
        \STATE Traverse parents from ${u^*}^{0}$ for the most relevant path $\pi*$
        \RETURN $\pi^*$
    \end{algorithmic}
\end{algorithm}

\subsection{Proof of Proposition \ref{prop:goat-lrp}}

\begin{proof}
    We expand \eqref{eq:gnn-gi} by writing
    \begin{equation}
        \partials[(x_{j_\ell})_{k_\ell}]{(x_{j_{\ell-1}})_{k_{\ell-1}}} = P^{(\ell)}_{j_\ell, k_\ell} \omega_{j_{\ell-1},j_\ell}^{(\ell)} W^{(\ell)}_{k_{\ell-1},k_\ell}.
    \end{equation}
    Upon noticing that
    \begin{equation}
        \omega_{j_{\ell-1},j_\ell}^{(\ell)} = A^{(\ell)}_{j_{\ell-1},j_\ell}
    \end{equation}
    we obtain \eqref{eq:goat-expansion}, completing the proof.
\end{proof}

%% file: appendix/accuracy_table.tex
\begin{tabular}{|l|ccccc|ccccc|ccccc|}
\hline
         & \multicolumn{5}{c}{GCN}               & \multicolumn{5}{|c|}{GAT}               & \multicolumn{5}{c|}{GIN}               \\
         \hline
\textbf{Layers} & 1     & 2     & 3     & 4     & 5     & 1     & 2     & 3     & 4     & 5     & 1     & 2     & 3     & 4     & 5     \\
         \hline
Cora     & 83.50 & 82.70 & 83.30 & 81.69 & 82.90 & 83.70 & 82.70 & 81.49 & 82.09 & 83.30 & 82.70 & 82.70 & 79.88 & 81.49 & 82.49 \\
Citeseer & 68.62 & 68.92 & 67.72 & 67.27 & 69.07 & 69.67 & 70.57 & 69.82 & 70.27 & 68.32 & 69.97 & 67.87 & 67.27 & 67.87 & 67.57 \\
Cornell  & 48.64 & 48.64 & 37.83 & 37.83 & 48.64 & 48.64 & 48.64 & 54.05 & 37.83 & 40.54 & 43.24 & 48.64 & 48.64 & 43.24 & 43.24 \\
Texas    & 45.94 & 32.43 & 37.83 & 32.43 & 43.24 & 54.05 & 35.13 & 48.64 & 27.03 & 64.86 & 32.43 & 24.32 & 37.84 & 35.14 & 40.54 \\
Wisconsin & 49.01 & 43.13 & 39.21 & 43.13 & 43.13 & 60.78 & 50.98 & 39.22 & 45.10 & 56.86 & 45.10 & 33.33 & 27.45 & 29.41 & 41.18 \\
\hline
MUTAG    & 69.23 & 63.46 & 73.08 & 73.08 & 73.08 & 73.08 & 65.38 & 65.38 & 65.38 & 61.54 & 84.62 & 76.92 & 88.46 & 76.92 & 73.08 \\
PROTEINS & 55.78 & 64.41 & 69.52 & 70.09 & 67.11 & 64.44 & 72.56 & 71.09 & 58.69 & 50.00 & 59.29 & 75.03 & 73.06 & 73.09 & 74.70 \\
Graph-SST2 & 80.76 & 83.64 & 87.38 & 88.19 & 85.29 & 83.35 & 81.89 & 84.84 & 50.00 & 50.00 & 82.13 & 87.64 & 87.43 & 88.30 & 88.19 \\ \hline
\end{tabular}

%% file: appendix/regularization.tex
\section{Analysis of Regularization Terms in GNNExplainer}
\label{appdx:regularization}

Here we analyze the effect of the regularization terms $\alpha\norm{\Omega}_1$ and $\beta H(\Omega)$ in GNNExplainer's optimization objective \eqref{eq:gnnexplainer}. We begin by analyzing the size term $\alpha\norm{\Omega}$, which is straightforward. For each edge weight $\omega_{ij}$ we have
\begin{equation}
    \partials{\omega_{ij}} \alpha\norm{\Omega}_1
    = \partials{\omega_{ij}} \alpha \sum_{(u,u') \in E} \omega_{ij}
    = \alpha
\end{equation}
for any value of $\omega_{ij}$, since each $\omega_{u,u'} \geq 0$.

To analyze the entropy term, we first briefly discuss the initialization strategy for GNNExplainer. Both the default parameters of the official implementation given by \cite{ying2019gnnexplainer} and the implementation given in PyTorch Geometric \cite{pyg} use Kaiming He initialization \cite{he2015delving} composed with a sigmoid function to map the result to $[0,1]$. Hence at $t = 0$, 
\begin{equation}
    \E\left[{\omega_{ij}}_{(t=0)}\right] = \frac{1}{2}.
\end{equation}
Then a standard calculation gives
\begin{equation}
    \partials{\omega_{ij}}H(\Omega) = -\frac{1}{|E|}\log\left(\frac{\omega_{ij}}{1 - \omega_{ij}}\right)
    \label{eq:entropy-derivative}
\end{equation}
so in expectation, we have at initialization that
\begin{equation}
    \E\left[\partials[H(\Omega)]{\omega_{ij}}\Big|_{\omega_{ij} = {\omega_{ij}}_{(t=0)}}\right] = 0.
\end{equation}
Thus, in expectation, the term $\beta H(\Omega)$ has no effect at initialization. However, if $\omega_{ij} > 1/2$, we see that \eqref{eq:entropy-derivative} becomes negative. Similarly, if $\omega_{ij} < 1/2$, \eqref{eq:entropy-derivative} becomes positive. Thus, in expectation, the effect of the entropy term will be determined by the first step of gradient descent, which is determined by
\begin{equation}
    - \eta\left(\partials{\omega_{ij}}\CE(Y_v,\Phi(v;G_S(\Omega))) + \alpha\right)
\end{equation}
where $\eta$ is the learning rate. If we have
\begin{equation}
    \partials{\omega_{ij}}\CE(Y_v,\Phi(v;G_S(\Omega))) + \alpha < 0
\end{equation}
then to minimize \eqref{eq:gnnexplainer}, GNNExplainer will take a step towards 1. In expectation, this means that ${\omega_{ij}}_{t=1} > 1/2$ and so 
\begin{equation}
    \partials{\omega_{ij}}H(\Omega)|_{\omega_{ij} = {\omega_{ij}}_{t=1}} = -\frac{1}{|E|}\log\left(\frac{{\omega_{ij}}_{t=1}}{1 - {\omega_{ij}}_{t=1}}\right) < 0.
\end{equation}
Similarly, if
\begin{equation}
    \partials{\omega_{ij}}\CE(Y_v,\Phi(v;G_S(\Omega))) + \alpha > 0
\end{equation}
then
\begin{equation}
    \partials{\omega_{ij}}H(\Omega)|_{\omega_{ij} = {\omega_{ij}}_{t=1}} > 0
\end{equation}
as well. Thus under the conditions analogous to \cref{prop:no-grad-flip}, where the expression
\begin{equation}
    \partials{\omega_{ij}}\CE(Y_v,\Phi(v;G_S(\Omega))) + \alpha
\end{equation}
is the same sign for all values of $\omega_{ij}$, the entropy regularization merely accelerates the gradient descent towards 0 or 1.

%% file: appendix/training.tex
\section{Training Details}
\label{appdx:training}

We conducted our experiments using PyTorch \cite{pytorch} and PyTorch Geometric \cite{pyg} on NVIDIA RTX A6000 GPUs.

\subsection{Synthetic Data}

For all synthetic experiments we use a GraphSAGE network \cite{hamilton2017inductive} with sum aggregation, ReLU activations, and no subsampling of node neighborhoods.

\paragraph{Negative Evidence}

We train a 1-layer network with no bias terms and no root weight using the Adam optimizer \cite{kingma2014adam} with a learning rate of $\eta = 0.01$ and $L_1$ regularization ($\gamma = 0.01$) over 1,000 epochs. Here $L_1$ regularization encourages the network to learn the ground truth. For GNNExplainer, we set the entropy regularization to $\beta = 0$, as all explanations will be connected. To more closely match the scenario in \cref{subsec:one-layer}, we use a smaller edge regularization of $\alpha = 0.001$ and a learning rate of $0.5$ to ensure GNNExplainer reaches convergence. The model reaches $100\%$ accuracy on the test set.

\paragraph{Infection}

We train a 4-layer sum-aggregated GraphSAGE with ReLU activation and a hidden layer width of 20. We use Adam for 100 epochs with a learning rate of $\eta = 0.005$ and weight decay ($L^2$ regularization) with $\gamma = 3\times 10^{-4}$. The model achieves perfect accuracy on the test graph. For GNNExplainer, we use the default parameters provided by PyTorch Geometric \cite{pyg}, training over 100 epochs with a learning rate of $\eta = 0.003$ and hyperparaeters $\alpha = 0.005$ and $\beta = 1$.

\subsection{Real Data}

Each GNN has a hidden width of 32 and is trained using Adam ($\eta = 0.003$) using a dropout of 0.5. For the GATs we use 4 attention heads and the GATv2 described in \cite{brody2021attentive}. For datasets with edge features, we augment GIN with the GINE layers from \cite{hu2019strategies}. For node classification (Cora, Citeseer, Cornell, Texas, Wisconsin) we train for 1000 epochs with weight decay of $\gamma = 10^{-5}$. For MUTAG and PROTEINS we train for 2000 epochs with weight decay of $\gamma = 10^{-4}$ and a batch size of 32. For Graph-SST2, we train for 100 epochs using a batch size of 128 and $\gamma = 10^{-6}$. Each graph classification model uses mean pooling to produce the final graph-level output, as well as layer normalization. \cref{tab:model-accuracy} shows the test performance of each model. For node classification we report accuracy. The graph classification tasks are binary, so we report AUROC.